# A Baseline Neural Machine Translation System for Indian Languages

**Jerin Philip*** IIIT Hyderabad

**Vinay P. Namboodiri**† IIT Kanpur

**C. V. Jawahar**‡ IIIT Hyderabad

## Abstract

We present a simple, yet effective, Neural Machine Translation system for Indian languages. We demonstrate the feasibility for multiple language pairs, and establish a strong baseline for further research.

## 1 Introduction

Ability to access information from non-native languages has become a necessity for humans in the modern age of Information Technology. This is specially true in India where there are many popular languages. While knowledge repositories in English are growing at rapid pace, Indian languages still remain very poor in digital resources. This leaves an average Indian handicapped in accessing the knowledge resources. Automatic machine translation is a promising tool to bridge the language barrier, and thereby to democratize the knowledge.

Machine Translation (MT) has been a topic of research for over two decades now, with schools of thought following rule-based approaches [Sinha et al., 1995, Chaudhury et al., 2010], example-based approaches [Somers, 1999, Sinhal and Gupta, 2014] and statistics-based approaches [Brown et al., 1993, Koehn, 2009, Kunchukuttan et al., 2014, Banerjee et al., 2018]. A class of statistics-based methods, involving the use of deep neural networks, popularly known as Neural Machine Translation (NMT) is making rapid progress in providing nearly usable translations across many language pairs [Edunov et al., 2018, Aharoni et al., 2019, Neubig and Hu, 2018]. Multiple competitive neural network architectures have come up in the last couple of years, each one outperforming the previous best and establishing superior baselines in a systematic manner [Sutskever et al., 2014, Bahdanau et al., 2014, Luong et al., 2015, Gehring et al., 2017, Vaswani et al., 2017]. Unfortunately, most of these methods are highly resource intensive. This, naturally made NMT less attractive for Indian languages due to the concerns of feasibility. However, we believe that the neural machine translation schemes are most appropriate for the Indian languages in the present day setting due to (i) the simplicity of the solution in rapidly prototyping and establishing practically effective systems, (ii) the lower cost of annotation of the training data. (The data required for building NMT systems at most demand sentence level alignments, with no special tagging.) (iii) the ease in transfer of ideas/algorithms across languages (Many practical tricks of developing effective solutions in one language provide insights in newer language pairs also.) (iv) ease in transfer of knowledge across languages/tasks, often implemented through pre-training, transfer learning or domain adaptation (v) rich software infrastructure available for rapidly prototyping effective models (eg. software stack for training) and deployment (eg. on embedded devices). Many of these are very important in the Indian setting, where the number of researchers and industries interested in Indian Language MT is very small compared to what is demanded. Often we may not even have more than one group equipped per language, with the state of the art tools and techniques.

India has 22 official languages, and many unofficial languages in use. With the fast growing numbers of mobile phone and Internet users, there is an immediate need for automatic machine translation systems from/to English as well as, across Indian languages. Though the digital content in Indian languages has increased a lot in the last few years, it is not yet comparable to that in English.

* jerin.philip@research.iiit.ac.in
† vinaypn@iitk.ac.in
‡ jawahar@iiit.ac.in

For example, there are only 780K Wikipedia pages in Hindi and 214K in Telugu in contrast to 36.5M in English [Wikipedia]. The ability to have effective translation models with acceptable performance takes solutions based on natural interfaces to the remote areas of a large world population. India has a spectrum of languages which could be labelled as low resource. Within the scope of the discussion in this work, a high resource language is one where large number ($> 1M$) of sentence-level aligned text pairs between two languages are available. Note that this itself could be a very conservative definition.

Usable NMT solutions already exist between several high resource languages of the world. Some closed systems are available for public use on the web [Google, Microsoft] for Indian Languages. However, they are limited in terms of details, reproducibility, and scope for further extensions to the wider research community.

The objective of this work is multi fold. We summarize them below.

- Though neural machine translation has received much attention for many western language pairs, not many effective systems are reported yet for Indic languages in the literature, barring exceptions like [Banerjee et al., 2018, Sen et al., 2018, Philip et al., 2018]. We would like to explore this machine learning and machine translation front, and demonstrate the feasibility for multiple Indian languages.

- Indian languages have always suffered from the lack of annotated corpora in the past for a number of language processing task including the machine translation that uses statistical or learning techniques. One of our objective is also to investigate how to effectively use the available data that could be monolingual, noisy and partially aligned.

- From a system point of view, we would like to build and demonstrate a web based system that can provide satisfactory results with possibly superior results to many of the previous open systems. In the process of developing a system, we also study the quirks that work for Indian languages, and comment on directions that we did not find promising.

- Quantitatively, we would like to establish a strong baseline for machine translation to and across Indian languages. We hope that this will enable further research in this area.

- Finally, we would like to release models and resources (such as noisy sentences) for enabling research in this area.

The rest of the content is organized as follows: We start off with Section 2 reviewing some of the relevant background literature with a specific focus on Indian language scenario. In Section 3, we discuss the components of the system and the adaptations carried out for Indian language situations. In Section 4, we elaborate the experimental details. In Section 5, we summarize the results and discuss our findings.

## 2 Machine Translation and Indian Language Situation

Machine translation systems could be built for a specific language pair or for multiple pairs simultaneously. We take the latter route in this work. This choice is dominated by many practical advantages in managing the limited available resources. We also hypothesize that the linguistic similairies across many Indian languages could help each other in this manner, though a systematic study on this is outside the scope of this work. Our work is in a multilingual setting wherein one system translates in all directions. We discuss challenges specific to the context of a few Indian languages.

**Text Resources** Most Indian language pairs are resource scarce in terms of sentence aligned parallel bi-text. More pairs and experts are likely to be available in pivoting through English to learn a multilingual translation model. Increasing reach of Internet access in the subcontinent is however changing these, as observable from Figure 1, which presents the systematic increase in the wikipedia pages in Indian languages.

Hindi-English (hi-en) can be labelled a relatively high resource pair, after continued efforts to compile and increase resources over the past decade. The IIT-Bombay Hindi English Parallel Corpus (IITB-hi-en) [Kunchukuttan et al., 2018] is the largest general domain dataset available for use in MT. In addition to a 1.5 Million parallel data, IITB-hi-en also offers nearly 19.2M monolingual hindi text, which can be used to further enhance the performance.

The Indian Languages Corpora Initiative (ILCI) Corpus [Jha, 2010] is another major initiative. The ILCI corpus we used here include parallel data specific to the domains of tourism and health across seven languages, all 7 aligned. Indic Languages Multilingual Parallel Corpus (WAT-ILMPC)[1] is a compilation of parallel data of subtitles, from OPUS[2], which contains parallel pairs of subtitles between languages of the subcontinent and English. We also make use of this.

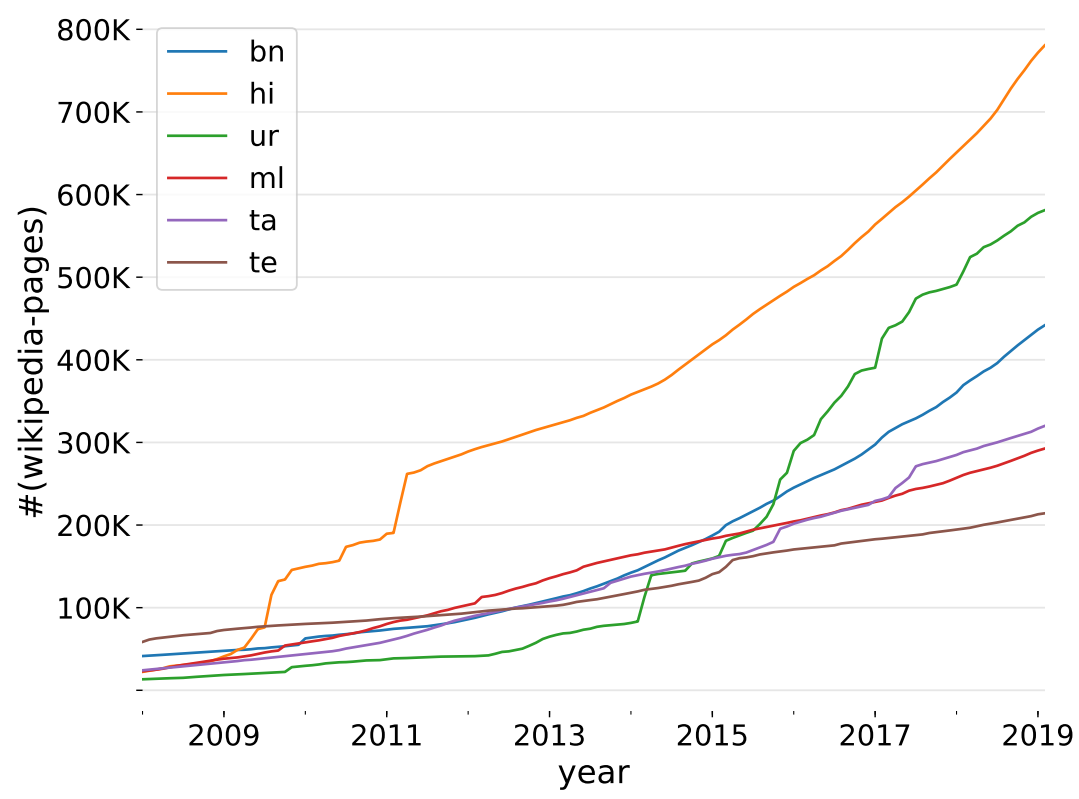

Figure 1: Growth of wikipedia pages in languages

**Linguistic Aspects** Khan et al. [2017] briefly covers linguistic aspects of several languages in the country. Indian languages are generally morphologically rich. Free word ordering in many languages lead to multiple ways of representing the content on the target side for some content in the source side. Dravidian languages face further challenges with heavy agglutinative and inflective characteristics. Recent developments in the use of open vocabulary [Kudo, 2018, Sennrich et al., 2015] have found success for many languages without large deterioration to translation quality. The same principles also help us in practically circumventing the challenges due to word morphology.

Our multilingual setting involves language pairs where translation is an "ill-posed" problem. For example, a single word "you" in English could translate to "tum", "me" and "aap" in Hindi, and valid sentences exist where correct translation can not be identified even with paragraph level contexts. Addressing these are left out of the scope of this paper.

[1] http://lotus.kuee.kyoto-u.ac.jp/WAT/indic-multilingual/index.html

[2] http://opus.nlpl.eu/

**Methods** Most of the past attempts look at the problem of machine translation in Indian setting as a pair-wise translation problem. That is, the scope is restricted to one or two languages to build translation solutions [Garje and Kharate, 2013, Khan et al., 2017]. The *Sampark* [TDIL-Sampark] system looks at translation between different pairs of languages in the country as multiple different problems, solved separately, with some commonality of the solution across the pairs in idea. This pairwise view of the problem space, enables greater incorporation of language expertise into the solution. In the absence of huge resources to train a statistical systems, these provided reasonable translations. *Sata Anuvadak* [Kunchukuttan et al., 2014], a compilation of 110 independently trained translation systems which used Statistical Machine Translation (SMT) analyzes a multilingual setup through SMT based approaches. One may attempt to solve multilingual translations through several hops through intermediate pivots. However, error compounds at each step.

Modern day NMT approaches are capable of handling sharing of learning between language pairs for which no data is available, through methods of zero shot learning [Johnson et al., 2017], transfer learning [Zoph et al., 2016], data-augmentation approaches [Sennrich et al., 2016, Edunov et al., 2018]. We are motivated by the success of these.

**Evaluations and Benchmarks** Workshop on Asian Translation (WAT 2018) [Nakazawa et al., 2018] tasks on Indian languages had seen models performing lesser on automated evaluation procedures being preferred more by human evaluation. This may lead to interesting explorations on the need of evaluation protocols and data sets for many Indian language translation tasks.

## 3 Neural Machine Translation

In the previous section, we discussed the design choices and the data we use in developing the model. We now discuss and formulate the basic building blocks of our NMT pipeline. We extend the formulation to our multilingual setup, and further discuss data augmentation techniques.

### 3.1 The Pipeline

We start with a joint vocabulary which can cover all source and target languages. The source and target samples can thus be represented as sequences comprising of a finite vocabulary $V$. We follow Kudo [2018] to reduce vocabulary to be feasible for learning, where the likelihood of a monolingual corpora is maximized while optimizing for minimum vocabulary.

Sequence to sequence models (seq2seq) with an encoder-decoder architecture are widely prevalent among NMT approaches [Sutskever et al., 2014, Bahdanau et al., 2014, Luong et al., 2015, Vaswani et al., 2017]. We also employ a similar architecture for translating text in one language to another. We now proceed to formally describe the parts relevant to this work below.

Consider a source sequence $\mathbf{x} = (x_1, x_2, \ldots, x_m)$ and a target sequence $\mathbf{y} = (y_1, y_2, \ldots, y_n)$, where $x_i, y_j \in V$. Our problem is to translate the sequence $\mathbf{x}$ to $\mathbf{y}$. Our translation model has an encoder ($f_\theta(\cdot)$) followed by a decoder. The encoder consumes the source side sequence $\mathbf{x}$ to learn representation $\mathbf{c}$. The auto-regressive decoder learns to maximize log-likelihood of the target sequence modelled by the conditional probability density given below. For a given time-step in the decoding process $t$, the generation process for the token at $t$ is conditioned on $\mathbf{c}$ and the tokens generated until $t-1$.

Encoding is represented as

$$\mathbf{c} = f_\theta(\mathbf{x})$$

Our objective is now to maximize

$$\sum_{t=1}^{t=T} \log p_\theta(y_t | y_{t'<t}, \mathbf{c})$$

at time/sequence index $t$. where

$$g(t) = \arg\max_{y_t \in V} \log p_\theta(y_t | y_{t'<t}, \mathbf{c})$$

All $(\mathbf{x}, \mathbf{y})$ pairs we have are mini-batched to fit onto resources available. The learnable parameter $\theta$ is optimized through mini-batch gradient descent of the objective. In this work, updates, steps or mini-batches are used synonymously. An epoch completes one visit of the entire training corpus. In our experiments, we use the Transformer [Vaswani et al., 2017] architecture. Transformer architectures powers most neural machine translation solutions offered today.

### 3.2 Multilingual Extensions

In a multilingual setting, we have several languages to translate to and from. Hereafter we use *xx*, *yy* and *zz* as placeholders to denote arbitrary languages. The languages and the respective notations (shown in the bracket) as English (en), Hindi (hn), Telugu (te), Tamil (ta), Malayalam (ml), Urdu (ur), Bangla (bn).

Our multilingual setup closely follows Johnson et al. [2017]. We share the encoder and decoder parameters across all languages, following the fact that the vocabulary is common on both sides and the representations being the same relaxes the computation. A control token is prepended to the input sequence to indicate which direction to translate to (`__t2xx__`). The decoder learns to generate the target given this input.

One advantage of using multilingual setups is that it enables zero shot learning. Few shot learning (such as zero-shot and one-shot) has emerged as an effective tool for addressing the lack of training data in many practical problems in computer vision [Socher et al., 2013, Zhang and Saligrama, 2015], language processing [Johnson et al., 2017] and speech processing [Stafylakis and Tzimiropoulos, 2018]. Pairs in language directions that did not exist in the training set can still be translated. They can also aid low resource settings of Indian languages. Rapid adaptations to new languages can also be done through multilingual setups [Neubig and Hu, 2018]. Aharoni et al. [2019] perform experiments at an even larger scale to find that massive-multilingual setups leading to large improvements in low resource setting.

### 3.3 Addressing Low Resource

In this work, we use a few, rather simple techniques to circumvent many challenges and demonstrate the utility for languages that are low-resource. These are briefly discussed.

**Transfer Learning** Zoph et al. [2016] demonstrate the success of transfer learning by using pairs of languages which have rich sentence level aligned bi-text in training low resource languages. A model is trained to take maximum advantage of the rich bi-text in for *zz*-*xx* pair, and the model instantiated with the weights from *zz*-*xx* is used to further train for *yy*-*zz*.

**Backtranslation** Using back-translation as a method to augment data used widely in NMT

methods today [Sennrich et al., 2015]. In this process, we train a model which can translate a low resource (LR) language to a high resource (HR) language. This model can attempt augmentation for training data by learning from a pair where it is high resource, followed by mapping the low resource languages to rich representations in high resource language. If there is sufficient monolingual data in the source side, this process creates rich parallel pairs. Edunov et al. [2018] take the idea further and tests at scale, to demonstrate improvements in translation.

## 4 Experimental Setup

In this section, we detail the datasets and the experimental setup used used in this work.

### 4.1 Datasets

In this work, we compile data available from several sources, each of which we describe below. The languages we use in our experiments include English (en), Hindi (hi), Bengali (bn), Malayalam (ml), Tamil (ta), Telugu (te) and Urdu (ur). Summary of the corpora used is provided in Table 1.

| | bn | hi | ml | ta | te | ur |
|---|---|---|---|---|---|---|
| wat-train | 337K | 84K | 359K | 26K | 22K | 26K |
| iitb-train | - | 1.5M | - | - | - | - |
| ilci-train | 50K | 50K | 50K | 50K | 50K | 50K |
| wat-dev | 0.5K | 0.5K | 0.5K | 0.5K | 0.5K | 0.5K |
| iitb-dev | - | 0.5K | - | - | - | - |
| iitb-test | - | 2.5K | - | - | - | - |
| wat-test | 1K | 1K | 1K | 1K | 1K | 1K |
| wat-mono | 453K | 105K | 402K | 30K | 24K | 29K |
| wiki-mono | 371K | - | 1M | 1.6M | 2.66M | - |
| iitb-mono | - | 19.2M | - | - | - | - |

Table 1: WAT Corpus Details

**Parallel Data** We primarily rely on the IITB-hi-en as parallel data for our training. We also use an extended version of our system from WAT-2018 [Philip et al., 2018] to backtranslate the available monolingual data in Hindi to English, and to augment the training.

To take full advantage in the multilingual setting, we use the ILCI corpus to include training pairs in all directions. In addition to the above, we use WAT-ILMPC, which gives ability to the multilingual model to implicitly pivot through English.

The designated splits for testing and validation (development) for IITB-hi-en and WAT-ILMPC are set aside for evaluations, while the remaining pool of data is used in training.

***Mann Ki Baat* test set** We extract a new multilingual test set for Indian languages, using data available online[3]. These are transcriptions of Prime Minister's address to the nation manually translated into multiple languages of the country. Our objective in creating this test-set is to find how well the models evaluated on WAT-ILMPC is able to generalize to a new data/situation.

**Monolingual Data** We use Wikipedia text to create samples to enhance model through back-translation. The IITB-hi-en and WAT-ILMPC come with their share of monolingual data as well, which we make good use of.

### 4.2 Training Details

In this section, we describe in detail the hyperparameters of the models in the previous section, in order to enable reproducibility.

We use `SentencePiece`[4] to reduce vocabulary while maintaing ability to cover the full text. While attempting to create one huge shared vocabulary building similar to Johnson et al. [2017], we find languages where pairs are lesser in number are under-represented. To get around this, we build separate SentencePiece models to deal with the class imbalance, and to take advantage of the large monolingual data available. These models are trained to cover maximum text constraining to 4000 vocabulary size. Due to shared vocabulary among these individual models arising from code-mixed content present in the large corpus, we end up with a vocabulary size of 26345, across 7 languages. One advantage to freezing the vocabulary is that we can perform warm starts for later models, while varying data.

Our experiments we use an NMT model based on Transformer-Base available in `fairseq-py`[5]. The model learns to translate across 7 languages. A control token to indicate which language to translate to is used to switch languages, as used in Johnson et al. [2017]. We train the model with all the available training data for 120 epochs, giving us a baseline model which we denote hereafter *IL-MULTI* . Computations for training *IL-MULTI*

[3] https://www.narendramodi.in/mann-ki-baat Thanks to *Bahubhashak* initiative for bringing the attention to this, as a language resource.

[4] https://github.com/google/sentencepiece

[5] https://github.com/pytorch/fairseq

took us approximately 3 days on 16 NVIDIA 1080Tis across 4 machines.

There are several issues with the data used to train *IL-MULTI* . One major concern visible is imbalance in results across languages. Given an NMT system which gives noisy but comprehensible translations to English, we use backtranslation to augment and bring in more authentic language samples for under-represented languages. Given monolingual corpus is abundant compared to bitext, this process works out to improve results further. Therefore we augment the training dataset with back-translations from monolingual corpus. We only add *{hi, en}->xx* samples, to deal with the class imbalance and to prevent decoder learning noisy text. The system we obtain after training *IL-MULTI* further for 72 epochs with backtranslated data added, we denote as *IL-MULTI+bt* .

We use `Bleualign`[6][Sennrich and Volk, 2011] to obtain parallel pairs between each pair of languages. Following this, we use heuristic methods to aligned pairs across languages, to obtain a near complete multilingual test-set. The approximated translation set for `Bleualign` is created by using our *IL-MULTI+bt* model, but the multilingual pairs after alignment undergoes one iteration of manual verification.

### 4.3 Evaluations

We report Bilingual Evaluation Understudy (BLEU) [Papineni et al., 2002] on all the test sets discussed previously. BLEU is a precision based metric for automatic evaluation of machine translation test sets. In Section 7, we describe and present other metrics which are used in the MT space to enable comparisons.

Automated evaluation is perhaps not the best choice when it comes to determine the quality of translations.

## 5 Results and Discussions

In this section, we present our results on WAT-ILMPC, IITB-hi-en and our newly explored *Mann Ki Baat* test-sets. In the process, we attempt to comprehensively study the performance.

### 5.1 Results on IITB-en-hi and WAT-ILMPC

We have now one-model yielding all-round performance on two tasks — one domain specific and one generic test set being trained on a combination of both datasets. We report the results on the automated evaluation procedures for the same here. A summary of our numbers on the BLEU evaluation metric for each task is presented in Table 2, across two variants of the same model we have.

Our base model i.e., *IL-MULTI* obtains the best results till date for the English to Hindi direction task for IITB-en-hi. The results for Hindi to English is also quite comparable.

For WAT-ILMPC tasks, our system gives comparable numbers while translating to English from other available languages. However, in the other direction, we observe that the results are lower. Attempts to fine-tune and adapt *IL-MULTI* to the domain specific subtitles dataset gives us *IL-MULTI+ft* , which demonstrates a boost in the evaluation scores on test sets specific to WAT-ILMPC. However, we observe performance degrades in the IITB-hi-en test set, which is general domain.

The overall ordering in the evaluation metrics show some patterns and consistency. While *xx-en* is giving comparable performances in many languages to the best values, *en-xx* fails to show similar performance. This could be due to imbalances in training samples among language-directions. The WAT-ILMPC dataset heavily pivots through English, the training data being strictly with English on one side and other languages on the other. This leaves training with a situation where there is a lot of data for the decoder to learn to generate English targets, while proportionately very less in languages of smaller resources. We try to solve this scenario by augmenting training with heavy monolingual data on the decoder, trying to translate from noisy backtranslated sources in English or Hindi. The model thus warm-started from *IL-MULTI* , *IL-MULTI+bt* gives us increase in BLEU for the less resource directions.

Our results on WAT-ILMPC and IITB-hi-en puts us in a comparable setting for many language pairs. However, our multiway training enables us to study how many non *xx-yy* directions work, where *xx* or *yy* need not necessarily be English. For this, we use the *Mann Ki Baat* dataset to obtain a proper multilingual dataset, using which we report our numbers in these directions below.

### 5.2 Multilingual Results

For our multilingual results, we present BLEU scores after using our *SentencePiece* models for tokenization. Though non-standard, this compen-sates for agglutinative languages with free word orderings ending up with low BLEU scores albeit translations being of reasonably good quality. Since the tokenization setup is uniform across languages, this lets us compare the several directions possible fairly. (We do however, report the non-*SentencePiece* tokenized BLEU scores in Section 7).

[6] https://github.com/rsennrich/Bleualign

| direction | model | wat-bn | wat-hi | wat-ml | wat-ta | wat-te | wat-ur | iitb-hi |
|---|---|---|---|---|---|---|---|---|
| from-en | *IL-MULTI* | 10.21 | 20.72 | 8.17 | 12.55 | 19.10 | 16.56 | 20.17 |
| | *IL-MULTI+ft* | 12.26 | 26.25 | 9.22 | 16.06 | 23.63 | 19.51 | 18.31 |
| to-en | *IL-MULTI* | 17.35 | 29.74 | 12.45 | 17.63 | 25.56 | 21.03 | **22.62** |
| | *IL-MULTI+ft* | 19.58 | 31.55 | 14.77 | 21.94 | 30.91 | 24.60 | 20.66 |

Table 2: BLEU scores on WAT and IITB test sets. Directions are indicated in the row.

| | |
|---|---|
| source | The Führer isn't recruiting you as soldiers, he's looking for a secretary. |
| target | यह आवश्यक नहीं है. फ्यूरर तुम को सैनिकों के रूप में भर्ती नहीं कर रहा है, वह एक सचिव खोज रहा है. |
| mm | फ्यूरर आप सैनिकों के रूप में भर्ती नहीं कर रहा है, वह एक सचिव की तलाश कर रहा है. |
| mm+ft | फ्यूरर आप सैनिकों के रूप में भर्ती नहीं कर रहा है, वह एक सचिव की तलाश में है। |
| source | No man in the sky intervened when I was a boy to deliver me from Daddy's fist and abominations. |
| target | आकाश में कोई भी आदमी हस्तक्षेप किया जब मैं एक लड़का था मेरे पिताजी की मुट्ठी और घिनौने काम से वितरित करने के लिए। |
| mm | आसमान में किसी आदमी ने तब हस्तक्षेप नहीं किया जब मैं पिताजी की मुट्ठी और घिनौनी से मुझे पहुंचाने वाला लड़का था। |
| mm+ft | आकाश में कोई भी आदमी हस्तक्षेप किया जब मैं एक लड़का था पिताजी की मुट्ठी और अपमान से मुझे देने के लिए. |
| source | Tonight, we offer you the one illusion that we dreamed about as kids but never dared perform until now. |
| target | आज रात, हम तुम हम बच्चों के रूप में के बारे में सपना देखा था कि एक भ्रम पेशकश... ... लेकिन अब तक प्रदर्शन करने की हिम्मत कभी नहीं. |
| mm | आज रात हम आपको एक भ्रम की पेशकश करते हैं कि हम बच्चों के रूप में सपने देखा लेकिन अब तक कभी हिम्मत नहीं की प्रदर्शन। |
| mm+ft | आज रात, हम तुम्हें एक भ्रम की पेशकश हम बच्चों के रूप में के बारे में सपना देखा कि लेकिन अब तक प्रदर्शन करने की हिम्मत कभी नहीं की. |
| source | What if I tell you it's got a Foot Locker on one side and a Claire's Accessories on the other. |
| target | क्या होगा अगर मैं तुम्हें बता मिल गया है एक फुट एक तरफ लॉकर और दूसरे पर एक क्लेयर सहायक उपकरण. |
| mm | क्या अगर मैं तुम्हें बताता हू कि यह एक तरफ एक फुट लॉकर और एक क्लेयर एक्सेसरीज मिल गया है. |
| mm+ft | क्या मैं यह एक तरफ एक फुट लॉकर मिल गया है और एक क्लेयर के एक्सेसरीज पर. |
| source | I could take up Jon's duties while he's gone, My Lord. |
| target | जबकि वह, चला गया मेरे प्रभु मैं जॉन के कर्तव्यों का समय लग सकता है। |
| mm | मैं जॉन के कर्तव्यों को संभाल सकता था जब वह गया है, मेरे प्रभु. |
| mm+ft | मैं जॉन के कर्तव्यों को संभाल सकता था जब वह चला गया है, मेरे प्रभु। |

Table 3: Five example sentences in English and their translations in Hindi. Readers may note the (i) length (ii) complexity and (iii) diversity among models of the sentences

In Table 4, we present results from *Mann Ki Baat*, using our model *IL-MULTI+bt* , which resulted the best numbers overall. The results indicate improvements especially in generating translations the resource scarce languages for a rich source. Our English-Hindi models give strongest performance, as in the setting above. The rows indicate source language and the columns target language.

From the grid, we see performance in (near) zero shot learning for language pairs which were not represented well enough in the dataset. The results in *ml-te* and *bn-te* show comparable performance with *hi-te* and *en-te*. The same phenomenon exists while translating to Tamil and Malayalam. Almost all the models have learned to copy well, i.e, when the source and target languages are same. It gets

| srcs | bn | en | hi | ml | ta | te | ur |
|---|---|---|---|---|---|---|---|
| bn | 99.87 | 15.20 | 14.43 | 8.15 | 3.85 | 7.35 | 0.0 |
| en | 9.50 | 100.00 | 15.37 | 8.71 | 4.60 | 8.20 | 0.0 |
| hi | 11.97 | 21.79 | 99.20 | 9.09 | 5.83 | 9.49 | 0.0 |
| ml | 6.85 | 12.00 | 9.84 | 99.90 | 3.37 | 7.32 | 0.0 |
| ta | 3.51 | 6.92 | 5.86 | 4.06 | 99.91 | 3.63 | 0.0 |
| te | 6.99 | 11.06 | 9.54 | 8.00 | 3.59 | 99.74 | 0.0 |
| ur | 0.00 | 0.00 | 19.55 | 0.00 | 0.00 | 0.00 | 100.0 |

Table 4: BLEU scores over translation pairs in Mann Ki Baat multilingual test-set. The complexity of sentences are high, with long and rich sentences of the kind included in ministerial speeches.

the target accurate $\sim$ 100% of the time.

In our experiments we find that *IL-MULTI+ft* which performed in the WAT-ILMPC test sets better failed to show good results here, indicating that the model is clearly adapting to WAT-ILMPC data.

### 5.3 Qualitative Examples and Discussions

In Tables 9 and 10, we illustrate the success and failure cases among a few languages.

Our model has managed to capture meaning most of the time. When the ground-truth is incorrect due to the subtitle corpus being noisy, our model yet manages to give the correct translation (WG -3 for example). There are samples like music tracks which are not translated and stay the same. While these are acceptable in a subtitle setting it is not so for pure translation attempts.

Sometimes the ground truth, being subtitles seem to have sentences for which word-meanings can not be inferred from the given context. Take WB-4 in Table 10 for example. The ground truth has a lot of context, assuming the person has a disease and is recovering while translating. It can be noticed that our models confuse it to be improvement or betterment, with the context being absent in the source.

Our model is not able to keep up to the mark while translating poetic sentences. It seems to be missing out on colloquial sentences as well, translating almost literally, missing intent although capturing enough meaning, as in WB-5.

## 6 Discussions

It is important to isolate errors or noise and improve the evaluation standards over time to enable more authentic evaluations. In the previous section, we illustrated several qualitative examples where our model was working fine, but the ground truth itself is incorrect, incomplete or has extra information. In this section, we take one iteration at improving the state of things of the test set. down a few errors and discuss our take on the evaluation metrics.

### 6.1 Shortcomings in the Evaluation Set

In our experiments, we found our multilingual models trained on large data to be robust and well performing for translations, but the BLEU scores obtained during evaluation indicates otherwise. Our best results are obtained after fine-tuning a general purpose model on the training set of the WAT-ILMPC. Further inspections indicate the ground truth samples to be incomplete, at either sides or both. In Table 5, we illustrate a few of these.

### 6.2 Limitations of the Evaluation Metrics/Protocol

In this section, we argue why several automated metrics or evaluation protocols may not be ideal for evaluating MT models in the Indian language setting. There are, like in many other languages, diverse ways of conveying the same meaning in text in Indian Languages as well. Normally, automated metrics like BLEU are evaluated across multiple references, but the WAT-ILMPC test set has only one reference.

## 7 Conclusion and Future Work

In this work, we report baselines for Indian Language tasks. Our solution implemented through the state of the art machine learning tricks yield competetive results for many language pairs even if not enough resources are available. We study the multilingual training scenario for Indian language, applying several established methods in literature successfully to get competitive results.

A reasonably functioning MT-System can be used in many other tasks. These involve computing embeddings for downstream tasks of classification or sentiment analysis. Ability to transfer the learning from well experimented embeddings like in English will help Indian Languages.

Some of the future works could include (i) use of ideas from machine learning to address the resource constraints (eg. few shot learning, active learning, adverserial training) in Indian language situations (ii) domain adaptation, transfer learning and style transfer in language/translation situations (iii) visualization and interpretebable translation models for Indian languages with sharable linguistic structures (iv) human in the loop solutions for training, inference and adaptations (v) Multimodal (image/video, speech and text) translational models and evaluation schemes.

## Acknowledgements

Authors would like to acknowledge (i) motivations from the consistent and focused efforts of Profs RS and VC over the last many decades and establishing research groups with clear focus (ii) the vision and thoughts of Prof. RR in empowering rural India, (iii) the efforts of Prof. PB in creating resources, test-beds and bringing international and systematic attention to the Indian language problems, and (iv) a large number of researchers who released their codes and resources publicly to enable this work. Special thanks to Prajwal Renukanad, Kiran Devraj, Vishnu Dorbala, Rudrabha, Binu, and other volunteers for helping with assessing the translation system for qualitative samples.

| | |
|---|---|
| source | Or you could leave and return to your families as men instead of murderers. |
| target | അല്ലെങ്കിൽ നിങ്ങൾക്ക് കൊലയാളികൾക്ക് പകരം നിൻറെ കുടുംബത്തിൽ മടങ്ങിപ്പോകാൻ കഴിയും. |
| source | I always wanted to build a talkies in our village in your father's name. |
| target | നിന്റെ അച്ഛന്റെ പേരിൽ നാട്ടിൽ ഒരു ടാക്കീസ് പണിയുക എന്നത് പണ്ട് മുതലേ ഉള്ള ഒരഗ്രഹമാണ്, മോനെ. |
| source | At the funfair, near the ghost train, the marshmallow twister is twisting. |
| target | मेले में, भूतीया train के पास, marshmallow घूमता ज रह है |
| source | See how lovingly he still stares at his master? |
| target | देखो कि वह अपने गुरु पर अब भी कैसे प्यार करता है? |
| source | And that's for Vortex |
| target | এবং সেটা কিনা ভরটেক্স'র হয়ে? পরিস্কার বুঝা যাচ্ছে তোমার কাছে সবার নাম আছে। |

Table 5: Qualitative sample from WAT test-dataset. Missing tokens are colored in red. Few samples which are treated as gold aren't as good.

| | |
|---|---|
| src.hi | मुंबई में उन्होंने भारतीय पारी की अंतिम गेंद पर आउट होने से पहले 15 गेंदों पर 26 रन बनाए। |
| general.en | In Mumbai, he made 26 runs on 15 balls before being out on the last ball of Indian mercury. |
| adapted.en | In Mumbai, he scored 26 runs in 15 balls before being dismissed for the last ball of the Indian innings. |
| src.hi | लेकिन लियोन ने ये कैच हवा में उड़कर लिया। |
| general.en | But Leon flew these coaches in the air. |
| adapted.en | But Leone took the catch in the air. |
| src.hi | 5- भारतीय कप्तान विराट कोहली श्रीलंका के खिलाफ वनडे मैचों में 2000 रन पूरे कर सकते हैं। |
| general.en | 5- Indian captain Virat Kohli can complete 2000 runs in forest matches against Sri Lanka. |
| adapted.en | 5- Indian captain Virat Kohli can complete 2000 runs in ODIs against Sri Lanka. |

Table 6: Translated samples by a model trained on a general dataset versus the same model adapted to a domain, here - cricket. The adapted samples are correct translations of the phrases. But due to mixing in English terms which are domain specific the general model makes errors.

# A Fine tuning to a specific target domain

Transfer learning has been widely used by the computer vision community for some time now. These works typically achieve excellent performance on a specific narrow domain task by fine-tuning a generic model trained on a larger dataset like ImageNet [Deng et al., 2009]. in the language space, past works attempt at fine tuning embeddings [**??**] for sentiment classification to a new language or domain where data is scarce, effectively transferring learning from the general case the embeddings are originally trained on. In this section, we experiment similar fine-tuning methods and show that we can improve our translation system for a particular domain – cricket related content.

Newspaper reports and commentary archives on cricket are crawled and corrected, and a noisy parallel pair is obtained using online services. 279K parallel pairs are thus obtained between Hindi and English. We set aside 100 samples as test-set and collect human provided translations for the same.

| Model | en → hi | hi → en |
|---|---|---|
| cold-start | 30.80 | 33.84 |
| warm-start | 25.57 | 29.27 |
| + fine-tuning | **35.72** | **39.97** |

Table 7: BLEU scores of different takes on building a domain specific translation system, for cricket. Cold-start denotes model trained on cricket data alone. Warm start begin with *IL-MULTI* - trained on the general dataset. This model is further fine tuned to cricket data.

We use three types of models while testing - (1) *IL-MULTI* , (2) *IL-MULTI* fine-tuned on cricket data and (3) a model with same parameters as *IL-MULTI* trained from scratch. Table 7 compares the performance of the three models on the test set. Among the three models, the warm-started model fine tuned on cricket specific corpus achieves the best performance, outperforming the model trained from scratch on this narrow domain.

Cricket domain include many terms which resolve to imply meanings different from their normal ones, when restricted to the domain. For instance a “boundary” or “four” would mean the event of “four runs” being scored in a match. Terms like “sixers”, “square drive”, “form” etc pose different meanings compared to the meanings associated with the same when placed in a general context. In Table 6 are some examples corrected by the fine-tuning method, which would otherwise be mistranslations due to the lack of contextual knowledge.

# B Additional Results

## B.1 More on WAT-ILMPC

In the main paper, the content was focused on building a general purpose machine translation system among 7 languages. We argued in favour of keeping *IL-MULTI* for future experiments against *IL-MULTI+ft* . In this section, we present supplementary content assessing *IL-MULTI+ft* , specific to the merits in particular to translation tasks WAT-ILMPC.

In Table 8, we indicate Rank-based Intuitive Bilingual Evaluation Score (RIBES)[Isozaki et al., 2010] and Adequacy Metric-Fluency Metric (AM-FM)[Banchs et al., 2015] additionally, along with the BLEU scores presented before. These numbers are evaluated by the WAT Submission platform and are comparable to other submissions.

| metric | direction | wat-bn | wat-hi | wat-ml | wat-ta | wat-te | wat-ur |
|---|---|---|---|---|---|---|---|
| BLEU | from-en | 12.26 | 26.25 | 9.22 | 16.06 | 23.63 | 19.51 |
| | to-en | 19.58 | 31.55 | 14.77 | 21.94 | 30.91 | 24.60 |
| RIBES | from-en | **0.654716** | 0.740006 | **0.517447** | 0.700706 | 0.779194 | 0.657035 |
| | to-en | **0.759463** | **0.805115** | 0.727629 | **0.770894** | **0.816129** | 0.741251 |
| AM-FM | from-en | **0.576480** | 0.652810 | **0.645120** | 0.760910 | 0.731550 | 0.537590 |
| | to-en | **0.529720** | **0.633010** | 0.521020 | 0.602280 | **0.667160** | 0.566190 |

Table 8: All metrics (BLEU, RIBES, AM-FM) on WAT-ILMPC leaderboard for our *IL-MULTI+ft* model. Our model ranks first in 11 cases, which are highlighted in bold.

Tables 9 and 10, we contains success and failure cases of our models on WAT-ILMPC test-sets, already discussed before.

## B.2 Mann Ki Baat

We present our multilingual translation BLEU scores on *Mann Ki Baat* test-set. We find what is expected in agreement with existing NMT literature.

Our baseline model gives us good results on the *Mann Ki Baat* test set. However, contrary to the case in WAT-ILMPC, we find that *IL-MULTI* outperforms *IL-MULTI+ft* . This is expected as we fine tuned to a subtitles corpus, and got better off results. As widely demonstrated in literature, augmentation with backtranslated data gives improvements of few BLEU points to models. Overall *IL-MULTI+bt* performs the best, as visible from the grids given below (Tables 11, 13, 12) [7].

From a qualitative point of view, we found samples from languages like *ml*, *te*, *ta* to be at par with languages except *hi* and *en*. However, the BLEU scores indicate a larger gap. We observe this is primarily due to agglutinative and inflective nature of these languages. Free word ordering in many leads to low BLEU scores, even when samples are of good quality.

The above numbers conform to normal tokenization in evaluation scripts primarily suited for English. But it is a common practice to use more information to tokenize. These will provide BLEU scores comparable to across languages, especially for languages with heavy agglutination and inflection. We present BLEU scores for the references and hypothesis tokenized using our SentencePiece models for the respective languages below in Tables 14, 16, 15.

[7] We are following indicnlp's BLEU computation, which should match the WAT website's BLEU computation, but did not. We are in touch with the organizers to make the values consistent.

| | | |
|---|---|---|
| WG -1 | wat.ml | അവർ ഒരു രോഗകാരിയെ പടർത്തുന്നതായി ഞാൻ കരുതുന്നു, അവർക്ക് ആരോഗ്യകരമായ ഒരു ഹോസ്റ്റ് വേണം. |
| | wat.en | I think they’re spreading a pathogen, and they need a healthy host. |
| | mm.en | I think they’re spreading a sick person, they need a healthy host. |
| | mm+ft.en | I think they’re spreading a patient, they need a healthy host. |
| WG -2 | wat.en | One, two, three, four, five, six, seven... eight, nine, ten, eleven. |
| | wat.ml | ഒന്ന്, രണ്ട്, മൂന്ന്, നാല്, അഞ്ച്, ആറു, ഏഴ്... എട്ട്, ഒമ്പത്, പത്ത്, പതിനൊന്ന്. |
| | mm.ml | ഒന്ന്, രണ്ട്, മൂന്ന്, നാല്, അഞ്ച്, ആറ്, ഏഴ്... 8, 9, 10, പതിനൊന്ന്. |
| | mm+ft.ml | ഒന്ന്, രണ്ട്, മൂന്ന്, നാല്, അഞ്ച്, ആറ്, ഏഴ്... എട്ട്, ഒമ്പത്, പത്ത്, പതിനൊന്ന്. |
| WG -3 | wat.en | Come on listen to this corporal for a moment sir if he’s not dead already |
| | wat.ml | Come on. സർജെന്റ്റ് , ഇയാൾ പറയുന്നത് ഒന്ന് കേൾക്കൂ. സർജെന്റ്റ്, ഇവിടെ മാരകമായി പരുക്കേറ്റ ഒരു പട്ടാളക്കാരനെയും കൊണ്ട് വന്നതാ. |
| | mm.ml | ഒരു നിമിഷം ഈ കോർപ്പറൽ പറയുന്നത് കേൾക്കൂ സർ, അവൻ മരിച്ചിട്ടില്ലെങ്കിൽ |
| | mm+ft.ml | ഈ കോർപ്പറൽ ഒരു നിമിഷം കേൾക്കൂ സർ... ...അവൻ മരിച്ചിട്ടില്ലെങ്കിൽ |
| WG -4 | wat.hi | तुम मेरे घर क्यों आये थे? |
| | wat.en | Then why do I hurt so bad right now? |
| | mm.en | Why did you come to my house |
| | mm+ft.en | Why did you come to my house |
| WG -5 | wat.hi | मुझे लगता है कि यह वास्तव में किसी तरह की एक उड़ान बात थी. |
| | wat.en | I think it was actually a flying thing of some kind. |
| | mm.en | I think it was actually a kind of flying thing. |
| | mm+ft.en | I think it was actually a kind of flying thing. |
| WG -6 | wat.hi | 16 दिसम्बर को हमारा काम कुछ-कुछ घुड़सवार सेना जैसा होगा, ओके? |
| | wat.en | Now, our role on the 16th of December is to be a bit like the cavalry, OK? |
| | mm.en | On December 16, our work will be like some horsemen army, OK |
| | mm+ft.en | On December 16th, our job will be like some riding army, okay |
| WG -7 | wat.en | I don’t know how you find the time to raise kids and teach. |
| | wat.ta | குழைந்தைக்கு கத்துகொடுக்குரதுக்கு எப்டி உங்களுக்கு நேரம் கெடச்சுது? |
| | mm.ta | நீங்கள் குழந்தைகள் வளர்க்கும் மற்றும் படிக்கும் நேரம் எப்படி தெரியாது. |
| | mm+ft.ta | நீங்கள் குழந்தைகள் வளர்க்கும் மற்றும் படிக்க நேரம் கண்டுபிடிக்க எப்படி எனக்கு தெரியாது. |

Table 9: Success Cases from WAT test-set.

| | | |
|---|---|---|
| WB-1 | wat.ml | അത് ഇനി നിൻറെ ഹൃദയത്തിൽ എത്ര വെറുപ്പും ദേഷ്യവും ഉണ്ടെന്ന് പറഞ്ഞാലും ശരി.. കാഞ്ചി വലിക്കാനുള്ള സമയമാകുമ്പോൾ.. മിക്ക ആൾക്കാർക്കും അതിന് കഴിയില്ല.. |
| | wat.en | No matter how much hate and anger you may have in your heart, when it comes time to pull the trigger most people can't do it. |
| | google.en | No matter how much hatred and anger it is in your heart .. most people can not do it .. |
| | mm+ft.en | If you say it's hate and angry in your heart, right... when you smoke the trigger, most people can't do it. |
| | mm.en | If it's a hate and anger in your heart, then it's okay... when it's time to smoke the treasure, most people can't do it. |
| WB-2 | wat.ml | വാളിനേക്കാൾ മൂർച്ച് അതു ചലിപ്പിക്കുന്നവൻറെ ചിന്തയിലാണ് ഉണ്ടാകേണ്ടത്. |
| | wat.en | Sometimes a sharp mind is enough. |
| | google.en | It should be in the mind of the one who moves it over the sword and moves it. |
| | mm+ft.en | It's more important than the sword that's in the mind of a moving man. |
| | mm.en | It's supposed to be in the mind of the sword that drives. |
| WB-3 | wat.en | "Most likely they have died of cold and hunger - far away there in the middle of the forest." |
| | wat.ml | "Most likely they have died of cold and hunger - far away there in the middle of the forest." |
| | mm.ml | "അവർക്ക് തണുപ്പും വിശപ്പും മരണം സംഭവിക്കും... ...അവിടെ വനത്തിൻറെ മദ്ധ്യത്തിൽ ദൂരെ." |
| | mm+ft.ml | "അവർക്ക് തണുത്തും വിശപ്പും മൂലം മരണം സംഭവിക്കാം... ...അവിടെ കാട്ടിൻറെ നടുവിൽ ദൂരെ." |
| WB-4 | wat.en | "Most likely they have died of cold and hunger - far away there in the middle of the forest." |
| | wat.en | He told me you would get better. And you did. |
| | wat.ml | അവൻ പറഞ്ഞു അമ്മയ്ക്ക് വേഗം ഭേദമാകുമെന്ന് അതു മാറുകയും ചെയ്തല്ലോ. |
| | mm.ml | അവൻ എന്നോട് പറഞ്ഞു നിനക്ക് നല്ലത് കിട്ടുമെന്ന്. |
| | mm+ft.ml | അവൻ എന്നോട് പറഞ്ഞു നിനക്ക് സുഖമാവുമെന്ന്. |
| WB-5 | wat.en | Old age should burn and rave at close of day |
| | wat.ta | Old age should burn and rave at close of day |
| | mm.ta | வயதானவர்கள் நாள் முழுவதும் எரிக்க வேண்டும்... |
| | mm+ft.ta | பழைய வயது எரிக்க வேண்டும் மற்றும் நாள் நெருக்கமாக குழாய் வேண்டும் |

Table 10: Failure Cases from WAT test-set

| srcs | bn | en | hi | ml | ta | te | ur |
|---|---|---|---|---|---|---|---|
| bn | 99.85 | **11.56** | 12.42 | 2.01 | 1.20 | 2.30 | 0.00 |
| en | 4.98 | 100.00 | **13.75** | 1.87 | 1.28 | 1.81 | 18.52 |
| hi | 6.93 | 17.87 | 99.49 | 2.77 | 1.71 | 3.09 | 12.55 |
| ml | 3.19 | **8.93** | 8.79 | 99.77 | 0.95 | 2.43 | 0.00 |
| ta | 1.68 | 5.15 | 5.33 | 1.16 | 99.74 | 1.57 | 0.00 |
| te | 2.78 | 7.51 | 7.67 | 2.24 | 1.06 | 99.48 | 0.00 |
| ur | 0.00 | 0.00 | 32.64 | 0.00 | 0.00 | 0.00 | 100.00 |

Table 11: BLEU scores of *IL-MULTI*

| srcs | bn | en | hi | ml | ta | te | ur |
|---|---|---|---|---|---|---|---|
| bn | 99.85 | 11.28 | **12.46** | **2.49** | **1.30** | **2.72** | 0.00 |
| en | 5.18 | 100.00 | 13.66 | **2.78** | **1.46** | **2.81** | 0.00 |
| hi | **7.01** | **17.95** | 99.50 | **3.03** | **2.07** | **3.78** | 0.00 |
| ml | **3.45** | 8.86 | **8.98** | 99.77 | **1.04** | **3.08** | 0.00 |
| ta | **1.78** | **5.28** | **5.46** | **1.22** | 99.85 | **1.71** | 0.00 |
| te | **3.01** | **7.64** | **7.84** | **2.61** | **1.21** | 99.68 | 0.00 |
| ur | 0.00 | 0.00 | 32.64 | 0.00 | 0.00 | 0.00 | 100.00 |

Table 12: BLEU scores of *IL-MULTI+bt*

| srcs | bn | en | hi | ml | ta | te | ur |
|---|---|---|---|---|---|---|---|
| bn | 99.49 | 9.53 | 3.07 | 1.07 | 0.59 | 1.46 | 0.00 |
| en | 4.19 | 99.99 | 10.10 | 1.93 | 1.18 | 1.63 | 0.00 |
| hi | 4.54 | 13.28 | 99.42 | 1.50 | 0.91 | 1.36 | 18.44 |
| ml | 1.33 | 7.51 | 4.79 | 99.42 | 0.88 | 1.89 | 0.00 |
| ta | 0.98 | 4.24 | 3.21 | 0.92 | 99.53 | 0.93 | 0.00 |
| te | 2.06 | 5.79 | 3.15 | 1.41 | 0.47 | 99.10 | 0.00 |
| ur | 0.00 | 0.00 | 0.00 | 0.00 | 0.00 | 0.00 | 100.00 |

Table 13: BLEU scores of *IL-MULTI+ft*

| srcs | bn | en | hi | ml | ta | te | ur |
|---|---|---|---|---|---|---|---|
| bn | 99.87 | 15.32 | 14.22 | 7.35 | 3.51 | 6.40 | 0.00 |
| en | 9.45 | 100.00 | 15.46 | 7.99 | 3.80 | 6.02 | 14.53 |
| hi | 11.96 | 21.77 | 99.22 | 8.50 | 4.55 | 7.34 | 10.50 |
| ml | 6.54 | 12.13 | 9.85 | 99.89 | 3.05 | 6.19 | 0.00 |
| ta | 3.40 | 6.82 | 5.85 | 3.69 | 99.86 | 3.20 | 0.00 |
| te | 6.51 | 11.00 | 9.34 | 7.42 | 3.26 | 99.65 | 0.00 |
| ur | 0.00 | 0.00 | 28.54 | 0.00 | 0.00 | 0.00 | 100.00 |

Table 14: BLEU scores of *IL-MULTI* with tokenization

| srcs | bn | en | hi | ml | ta | te | ur |
|---|---|---|---|---|---|---|---|
| bn | 99.87 | 15.20 | 14.43 | 8.15 | 3.85 | 7.35 | 0.0 |
| en | 9.50 | 100.00 | 15.37 | 8.71 | 4.60 | 8.20 | 0.0 |
| hi | 11.97 | 21.79 | 99.20 | 9.09 | 5.83 | 9.49 | 0.0 |
| ml | 6.85 | 12.00 | 9.84 | 99.90 | 3.37 | 7.32 | 0.0 |
| ta | 3.51 | 6.92 | 5.86 | 4.06 | 99.91 | 3.63 | 0.0 |
| te | 6.99 | 11.06 | 9.54 | 8.00 | 3.59 | 99.74 | 0.0 |
| ur | 0.00 | 0.00 | 19.55 | 0.00 | 0.00 | 0.00 | 100.0 |

Table 15: BLEU scores of *IL-MULTI+bt* with tokenization

| srcs | bn | en | hi | ml | ta | te | ur |
|---|---|---|---|---|---|---|---|
| bn | 99.73 | 12.85 | 3.87 | 4.29 | 1.05 | 4.31 | 0.00 |
| en | 8.39 | 99.98 | 12.34 | 7.70 | 2.95 | 4.94 | 0.00 |
| hi | 7.52 | 17.33 | 99.17 | 5.57 | 2.07 | 3.44 | 15.67 |
| ml | 3.11 | 10.33 | 6.08 | 99.75 | 2.50 | 4.61 | 0.00 |
| ta | 1.79 | 5.71 | 3.91 | 3.13 | 99.77 | 1.84 | 0.00 |
| te | 4.96 | 8.67 | 4.82 | 5.23 | 0.87 | 99.51 | 0.00 |
| ur | 0.00 | 0.00 | 0.00 | 0.00 | 0.00 | 0.00 | 100.00 |

Table 16: BLEU scores of *IL-MULTI+ft* with tokenization

| | | |
|---|---|---|
| en | Mann Ki Baat, February 2019 My dear countrymen, Namaskar. | |
| en | Mann Ki Baat, February 2019 My dear countrymen, Namaskar. | Mann Ki Baat, February 2019 My dear countrymen, Namaskar. |
| hi | मान की बात, फरवरी 2019 मेरे प्रिय देशवासी, नमस्कार। | मन की बात, फरवरी 2019 मेरे प्यारे देशवासियो, नमस्कार |
| bn | কি বাট, ফেব্রুয়ারি ২০১৯ আমার প্রিয় দেশবাসী, নামসকার. | মন কি বাত, ফেব্রুয়ারি ২০১৯ আমার প্রিয় দেশবাসী, নমস্কার! |
| ta | மன் கி பாட், பிப்ரவரி 2019 என் டியர் நேச நாட்டு நமஸ்கர். | மனதின் குரல், பிப்ரவரி 2019 எனதருமை நாட்டுமக்களே, வணக்கம்! |
| en | In order to ensure that their fellow countrymen could sleep peacefully, these brave sons toiled relentlessly, day or night. | |
| en | In order to ensure that their fellow countrymen could sleep peacefully, these brave sons toiled relentlessly, day or night. | In order to ensure that their fellow countrymen could sleep peacefully, these brave sons toiled relentlessly, day or night. |
| ml | ദേശവാസികൾക്ക് സമാധാനത്തോടെ ഉറങ്ങാൻ കഴിയുമെന്ന് ഉറപ്പുവരുത്താൻ, ഈ ധീരപുത്രൻമാർക്ക് നിത്യമോ, രാത്രിയോ ഉണർന്നില്ല. | ജനങ്ങൾ സമാധാനത്തോടെ ഉറങ്ങാൻ, നമ്മുടെ ഈ വീരപുത്രന്മാർ രാത്രിയെ പകലാക്കി കാവൽ നിന്നു. |
| hi | ताकि यह सुनिश्चित किया जा सके कि उनके साथी देशवासी शांतिपूर्वक सो सकें, इन बहादुर बेटों ने अथक, दिन या रात बिता दी। | देशवासी चैन की नींद सो सकें, इसलिए, इन हमारे वीर सपूतों ने, रात-दिन एक करके रखा था |
| te | దేశస్థులు ప్రశాంతంగా నిద్రపోవాలని శ్రద్ధగా నిశ్చయించేందుకు ఈ సాహస కుమారులు ప్రశాంతంగా, పగలు లేదా రాత్రికి తోడ్పడ్డారు. | దేశప్రజలు ప్రశాంతంగా నిద్ర పోవడం కోసం ఈ వీరపుత్రులు తమ నిద్రాహారాలు మానుకుని మనల్ని రక్షించారు. |
| ta | நண்பர்கள் அமைதியாக தூங்க முடிந்தது என்பதை உறுதி செய்ய இந்த வீரப் மகன்கள் அமைதியாக, நாள் அல்லது இரவில் துறந்தனர். | நாட்டுமக்கள் நிம்மதியாக உறங்கவேண்டும் என்பதற்காக, நம்முடைய இந்த வீர மைந்தர்கள், இரவு பகல் எனப்பாராமல் தங்களை அர்ப்பணித்திருந்தார்கள். |
| bn | তাদের দেশবাসীরা শান্তিতে ঘুমাতে পারতে নিশ্চিত করার জন্য, এই সাহসী ছেলেরা নিঃসন্দেহে, দিন বা রাতে বিরক্ত হয়ে যায়। | দেশবাসী যাতে নিশ্চিন্ত ঘুমোতে পারেন সেদিকে খেয়াল রেখে এইসব বীর সুসন্তানরা দিন-রাত এক করে সজাগ দৃষ্টি রেখে চলতো। |
| te | పది రోజుల క్రితం భరతమాత తన వీర పుత్రులను కోల్పోయింది. | |
| en | Ten days ago, the Great Mother lost his heroic sons. | 10 days ago, Mother India had to bear the loss of many of her valiant sons. |
| ta | பத்து நாட்களுக்கு முன்னால் பரத்மாத தன்னுடைய வீரபுத்திரர்களை இழந்தார் . | 10 நாட்களுக்கு முன்பாக, பாரத அன்னை தன் வீர மைந்தர்களை இழந்திருக்கிறாள். |
| hi | दस दिन पहले भरत मा अपने वीर पुतलों को खो चुकी थी । | 10 दिन पूर्व, भारत-माता ने अपने वीर सपूतों को खो दिया |
| bn | দশ দিন আগে ভৃত্য নিজের বীরের পুত্রকে হারিয়ে ফেলেন। | দশ দিন আগে, ভারতমাতা তাঁর বীর সুসন্তানদের হারিয়েছেন। |
| te | పది రోజుల క్రితం భరతమాత తన వీర పుత్రులను కోల్పోయింది. | పది రోజుల క్రితం భరతమాత తన వీర పుత్రులను కోల్పోయింది. |
| ml | പത്തു ദിവസത്തിനു മുന്പ് ഭരതൻ തൻറെ വീരപുത്രന്മാരെ നഷ്ടപ്പെട്ടു . | 10 ദിവസം മുമ്പ് ഭാരതമാതാവിന് ധീരന്മാരായ പുത്രന്മാരെ നഷ്ടപ്പെട്ടു. |
| ml | സൈന്യം ഭീകരവാദികളെയും അവരെ സഹായിക്കുന്നവരേയും വേരോടെ ഇല്ലാതെയാക്കാൻ ദൃനിശ്ചയം ചെയ്തിരിക്കുകയാണ്. | |
| en | The army is determined to seize the terrorists and those who help them. | The Army has resolved to wipe out terrorists and their harbourers. |
| hi | सेना ने आतंकवादियों और उनकी मददगारों को जड़ से वंचित करने का संकल्प लिया है। | सेना ने आतंकवादियों और उनके मददगारों के समूल नाश का संकल्प ले लिया है |
| ta | இராணுவம் பயங்கரவாதிகள் மற்றும் அவர்களுக்கு உதவி செய்தவர்களை வேரிலிருந்து விடுவிக்க முயற்சி செய்துள்ளது . | தீவிரவாதிகளையும் அவர்களுக்குத் துணை செல்பவர்களையும் வேரோடு கெல்லி எறியும் உறுதிப்பாட்டை இராணுவத்தினர் மேற்கொண்டிருக்கிறார்கள். |
| ml | സൈന്യം ഭീകരവാദികളെയും അവരെ സഹായിക്കുന്നവരേയും വേരോടെ ഇല്ലാതെയാക്കാൻ ദൃനിശ്ചയം ചെയ്തിരിക്കുകയാണ്. | സൈന്യം ഭീകരവാദികളെയും അവരെ സഹായിക്കുന്നവരേയും വേരോടെ ഇല്ലാതെയാക്കാൻ ദൃനിശ്ചയം ചെയ്തിരിക്കുകയാണ്. |
| te | సైనిక దౌత్యవేత్తలు మరియు వారి సహాయం చేసేవారిని వేరువేరు నుండి తప్పించుకోవడానికి నిశ్చయించబడింది . | అవే మాటలు దేశ ధైర్యానికి బలాన్ని ఇచ్చాయి. |
| bn | সৈন্যরা সন্ত্রাসীদের এবং তাদের সাহায্যকারীদেরকে আলাদা করে রাখার জন্য প্রস্তুত করা হয়েছে। | সেনারা সন্ত্রাসবাদীদের এবং তাদের মদতকারীদের কীভাবে সমূলে ধ্বংস করা যায় তার সঙ্কল্প নিয়েছে। |

Table 17: Qualitative samples from Mann Ki Baat Dataset. Source is one large row. Translations to different languages are shown in two columns - first one is our prediction and second one the respective aligned ground truth.

| | | |
|---|---|---|
| en | Jamuna Tudu, famous nicknamed ’Lady Tarzan’ in Jharkhand, most valiantly took on the Timber Mafia and Naxalites, and not only saved the 50 hectares of forest but also inspired ten thousand women to unite and protect the trees and wildlife. | |
| en | Jamuna Tudu, famous nicknamed ’Lady Tarzan’ in Jharkhand, most valiantly took on the Timber Mafia and Naxalites, and not only saved the 50 hectares of forest but also inspired ten thousand women to unite and protect the trees and wildlife. | Jamuna Tudu, famous nicknamed ’Lady Tarzan’ in Jharkhand, most valiantly took on the Timber Mafia and Naxalites, and not only saved the 50 hectares of forest but also inspired ten thousand women to unite and protect the trees and wildlife. |
| te | జార్ఖండ్‌లో ప్రసిద్ధి చెందిన "లాడీ టార్జాన్" పేరుగల "జమునా టూడూ", అత్యధికంగా టింబర్ మాఫియా మరియు క్షిపణిదారులపై తీసుకున్నారు, కేవలం 50 హెక్టార్ల అడవులను మాత్రమే కాపాడలేదు, కానీ పదివేల మంది స్త్రీలు చెట్లను మరియు వన్యప్రాణిని సంరక్షించడానికి కూడా ప్రేరణ చేశారు. | ◌ార్ఖండ్ లో లేడీ టార్జాన్ పేరుతో ప్రఖ్యాతి చెందిన జమునా టూడూ, టింబర్ మాఫియా తోనూ, నక్సలైట్ల తోనూ పోరాడే సాహసవంతమైన పని చేసారు. ఏభై హెక్టార్ల అటవీ ప్రాంతాన్ని నష్టపోకుండా కాపాడడమే కాకుండా, కలిసికట్టుగా చెట్లు, వన్యజీవాల రక్షణ కోసం పోరాడేలా పదివేల మహిళలకు ప్రేరణను ఇచ్చారు. |
| ta | ஜார்க்கண்டில் புகழ்பெற்ற "லாடி தர்ஜன்" ("Jamuna Tudu", "Timber Mafia" மற்றும் "Naki") மீது மிகவும் பிரபலமாக எடுத்தார், மேலும் 50 ஹெக்டேர் காடுகளை மட்டுமே காப்பாற்றியதுடன், பத்து ஆயிரம் பெண்களை மரங்களையும், வனவிலங்குகளையும் ஒருமைப்படுத்தவும் தூண்டினார். | நூறு ஆண்டுகள் ஆன போதிலும் கூட உலகம் முழுவதிலும் மக்களுக்கு யோகக்கலையை அவர் பயிற்றுவிக்கிறார், இதுவரை 1500 பேர்களை யோகப் பயிற்றுநர்களாக ஆக்கியிருக்கிறார். |
| ml | ഝാർഖണ്ഡിലെ 'ലാഡി ടാർസൻ' എന്ന പ്രശസ്തനായ ജമുന ടുഡു, ഏറ്റവും വിശാലമായി ടിംബർ മാഫിയയും ദേശാടനക്കാരും സഞ്ചരിച്ച് 50 ഹെക്ടർ വനം രക്ഷിച്ച മാത്രമല്ല, പത്ത് ആയിരം സ്ത്രീകൾ മരങ്ങളും വന്യജീവികളെയും ഒന്നിച്ച രക്ഷിക്കാനും പ്രേരിപ്പിച്ചു. | ◌ാർഖണ്ഡിൽ ലേഡി ടാർസൻ എന്ന പേരിൽ വിഖ്യാതയായ യമുനാ ടുഡൂ തടി മാഫിയയോടും നക്സലുകളോടും പോരാടുകയെന്ന സാഹസം പ്രവർത്തിച്ചു. 50 ഹെക്ടർ വനം നശിക്കാതെ കാത്തുവെന്ന മാത്രമല്ല, പതിനായിരം സ്ത്രീകളെ സംഘടിപ്പിച്ച് വൃക്ഷങ്ങളുടെയും വന്യജീവികളുടെയും രക്ഷക്കായി പ്രവർത്തിക്കാൻ അവരെ പ്രേരിപ്പിച്ചു. |
| bn | ঝাড়খন্ডে বিখ্যাত "লাডি টার্জন", সবচেয়ে চম□কারভাবে টিম্বার মাফিয়া ও নক্ষত্রবাহিনী দখল করেন। শুধুমাত্র ৫০ হেক্টর বন রক্ষা করেন, এমনকি দশ হাজার মহিলাকে গাছ ও বন্য প্রাণীকে একত্র করার জন্যও প্রেরণ করেছেন।. | ঝাড়খণ্ডের বিখ্যাত □Lady Tarzan□ হিসেবে খ্যাত যমুনা টুডু, গাছপাচারকারী মাফিয়া ও নকশালদের মোকাবিলা করার মত সাহসী কাজ করে তিনি শুধু ৫০ হেক্টর জঙ্গল উজাড় হয়ে যাওয়ার হাত থেকে বাঁচাননি, উপরন্তু দশ হাজার মহিলাকে একজোট করে গাছ ও বন্যপ্রাণীদের সুরক্ষার জন্য পাঠিয়েছেন। |
| hi | झारखंड में मशहूर उपनाम 'लेडी तर्जन' जमुना तुडू ने सबसे ज्यादा चमत्कार से टिंबर माफिया और नक्सलियों को न केवल 50 हेक्टेयर जंगल से बचाया बल्कि दस हजार महिलाओं को पेड़-पौधों और वन्यजीवों की सुरक्षा के लिए प्रेरित किया। | झारखण्ड में □□□ □□□□□ के नाम से विख्यात जमुना टुडू ने टिम्बर माफिया और नक्सलियों से लोहा लेने का साहसिक काम किया उन्होंने न केवल 50 हेक्टेयर जंगल को उजड़ने से बचाया बल्कि दस हज़ार महिलाओं को एकजुट कर पेड़ों और वन्यजीवों की सुरक्षा के लिए प्रेरित किया |

Table 18: Longer sentences from Mann Ki Baat set.